\documentclass{article}

\usepackage{microtype}
\usepackage{graphicx}
\usepackage{subfigure}
\usepackage{booktabs} 
\usepackage{enumitem}
\usepackage{wrapfig}
\usepackage[bottom]{footmisc}
\usepackage{amsmath, amsthm, amssymb}
\setlist[enumerate]{itemsep=0mm}

\usepackage[dvipsnames]{xcolor}
\newcommand{\tabincell}[2]{\begin{tabular}{@{}#1@{}}#2\end{tabular}}

\newcommand{\xingyu}[1]{#1}

\usepackage{hyperref}

\usepackage[final]{corl_2020} 

\title{SoftGym: Benchmarking Deep Reinforcement Learning for Deformable Object Manipulation}

\author{
  Xingyu Lin, ~Yufei Wang, ~Jake Olkin,~David Held \\
  Carnegie Mellon University \\
  United States\\
  \texttt{\{xlin3, yufeiw2, jolkin, dheld\}@andrew.cmu.edu} \\
}

\begin{document}
\maketitle


\begin{abstract}
Manipulating deformable objects has long been a challenge in robotics due to its high dimensional state representation and complex dynamics. Recent success in deep reinforcement learning provides a promising direction for learning to manipulate deformable objects with data driven methods. However, existing reinforcement learning benchmarks only cover tasks with direct state observability and simple low-dimensional dynamics or with relatively simple image-based environments, such as those with rigid objects. In this paper, we present SoftGym, a set of open-source simulated benchmarks for manipulating deformable objects, with a standard OpenAI Gym API and a Python interface for creating new environments. Our benchmark will enable reproducible research in this important area. Further, we evaluate a variety of algorithms on these tasks and highlight challenges for reinforcement learning algorithms, including dealing with a state representation that has a high intrinsic dimensionality and is partially observable. The experiments and analysis indicate the strengths and limitations of existing methods in the context of deformable object manipulation that can help point the way forward for future methods development.
Code and videos of the learned policies can be found on our project website.\footnote{\url{https://sites.google.com/view/softgym}}
\end{abstract}




\section{Introduction}
Robotic manipulation of deformable objects has wide application both in our daily lives, such as folding laundry and making food, and in industrial applications, such as packaging or handling cables. However, programming a robot to perform these tasks has long been a challenge in robotics due to the high dimensional state representation and complex dynamics~\cite{maitin2010cloth,miller2011parametrized,miller2012geometric}. 

One potential approach to enable a robot to perform these manipulation tasks is with deep reinforcement learning (RL), which has achieved many successes in recent years~\cite{mnih2015human,vinyals2019grandmaster,schrittwieser2019mastering,hwangbo2019learning,andrychowicz2020learning}. 
Some recent works have used learning-based methods to explore the challenges of deformable object manipulation~\cite{li2018learning,matas2018sim,agrawal2016learning,sermanet2018time}; however, these works often each evaluate on a different task variant with different simulators or robot setups, making it challenging to directly compare these approaches. There is currently no benchmark for evaluating and comparing different approaches for deformable object manipulation.

In contrast, there are a number of popular benchmarks for reinforcement learning with rigid or articulated objects~\cite{brockman2016openai,tassa2018deepmind,fan2018surreal}.  Many of these benchmarks assume that the agent directly observes a low dimension state representation that fully describes the underlying dynamics of the environment, such as the joint angles and velocities of the robot~\cite{brockman2016openai,duan2016benchmarking} or object state~\cite{andrychowicz2020learning,johannink2019residual}. However, such low-dimensional sufficient state representations are difficult to perceive (or sometimes even define) for many deformable object tasks, such as laundry folding or dough manipulation.  For deformable object manipulation, the robot must operate directly on its observations, which can include camera images and other sensors.

In this paper, we present SoftGym, a set of open-source simulated benchmarks for manipulating deformable objects, with a standard OpenAI Gym API and Python interface for creating new environments. Currently, SoftGym includes 10 challenging environments involving manipulation of rope, cloth and fluid of variable properties, with different options for the state and action spaces. These environments highlight the difficulty in performing robot manipulation tasks in environments that have complex visual observations with partial observability and an inherently high dimensional underlying state representation for the dynamics.  SoftGym provides a standardized set of environments that can be used to develop and compare new algorithms for deformable object manipulation, thus enabling fair comparisons and thereby faster progress in this domain.


\xingyu{We benchmark a range of algorithms on these environments assuming different observation spaces for the policy, including full knowledge of the ground-truth state of the deformable object, a low-dimension state representation, and only visual observations. Our results show that learning with visual observations leads to much worse performance compared to learning with ground-truth state observations in many deformable object manipulation tasks.} The poor performance of image-based methods on these environments motivates the need for future algorithmic development in this area.  We also provide an analysis to give some insight into why current methods that use visual observations might have suboptimal performance; this analysis can hopefully point the way towards better methods for deformable object manipulation.


\section{Related Works}
Robotic manipulation of deformable objects has a rich history across various fields, such as folding laundry~\cite{maitin2010cloth}, preparing food~\cite{bollini2013interpreting}, or assistive dressing and feeding~\cite{chen2013robots,erickson2018deep}. Early works used traditional vision algorithms to detect key features such as edges and corners~\cite{maitin2010cloth,willimon2011model,triantafyllou2011vision}. Motion planning is then adopted along with analytical models of the deformable objects~\cite{saha2007manipulation,rodriguez2006obstacle}. However, these planning approaches often suffer from the large configuration space induced by the high degree of freedom of the deformable objects~\cite{essahbi2012soft}. We refer to \cite{sanchez2018robotic,khalil2010dexterous} for a more detailed survey on prior methods for robot manipulation of deformable objects. 

Recently, the success of deep learning  has garnered increased interest in learning to solve perception and manipulation tasks in an end-to-end framework~\cite{agrawal2016learning,sermanet2018time,ebert2018visual,pathak2018zero,yan2020self}. In cloth manipulation, recent work uses demonstrations to learn an image-based policy for cloth folding and draping~\cite{matas2018sim}. Other works learn a pick-and-place policy to spread a towel~\cite{seita2019deep,wu2019learning}. Due to the large number of samples required by reinforcement learning, as well as the difficulty in specifying a reward function, all these works start by training the policy in simulation and then transfer the policy to a real robot through domain randomization and dynamics randomization. However, these papers do not systematically compare different methods on a range of tasks.


Standard environments with benchmarks have played an important role in the RL community, such as the Arcade Learning environments~\cite{bellemare2013arcade} and the MuJoCo environments~\cite{duan2016benchmarking}. A variety of new environments have been created recently to benchmark reinforcement learning algorithms~\cite{fan2018surreal,osband2019behaviour,yu2019meta,james2019rlbench,lee2019ikea}. However, none of these benchmark environments incorporate deformable objects, and usually the full state of the system can be represented by a low-dimensional vector. \xingyu{Other recent environments built on top of the Nvidia PhysX simulator also have the ability to simulate of deformable objects~\cite{Xiang_2020_SAPIEN,4633} but do not include any tasks or assets for manipulating deformable objects.} As such, we believe that SoftGym would be a unique and valuable contribution to the reinforcement learning and robotics communities, by enabling new methods to be quickly evaluated and compared to previous approaches in a standardized and reproducible set of simulated environments.

\begin{figure*}[h]
    \centering
    \includegraphics[width=\textwidth]{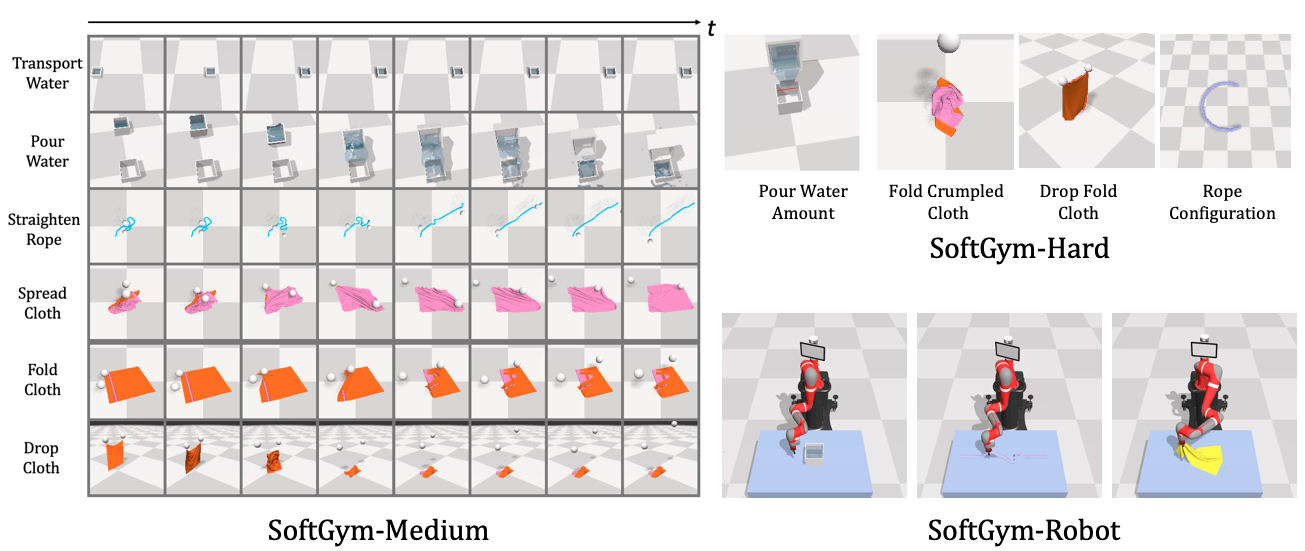}
    \vspace{-7mm}
    \caption{Visualizations of all tasks in SoftGym. These tasks can be used to evaluate how well an algorithm works on a variety of deformable object manipulation tasks.}
    \label{fig:task_visual}
\end{figure*}
\section{Background: Deformable Object Modeling in FleX}
SoftGym builds on top of  Nvidia FleX physics simulator. Nvidia's FleX simulator models deformable objects in a particle and position based dynamical system~\cite{muller2007position,macklin2014unified}. Each object is represented by a set of particles and the internal constraints among these particles. Each particle $p_i$ has at least three attributes: position $x_i$, velocity $v_i$, and inverse mass $w_i$. Different physical properties of the objects are characterized by the constraints. A constraint is represented in the form of $C(x_1, ..., x_n) \geq 0$ or $C(x_1, ..., x_n) = 0$, where $C(x)$ is a function of all the positions of the relevant particles. Given the current particle positions $p_i$ and velocities $v_i$, FleX first computes a predicted position $\hat{p}_i = p_i + \Delta t v_i$ by integrating the velocity. The predicted positions are then projected onto the feasible set given all the constraints to obtain the new positions of the particles in the next step.




Fluids can naturally be modeled in a particle system, as detailed in~\cite{macklin2013position}; specifically, a constant density constraint is applied to each particle to enforce the incompressibility of the fluid. For each particle, the constant density constraint is based on the position of that particle, as well as the positions of its neighboring particles.

Rope is modeled as a sequence of particles, where each pair of neighbouring particles are connected by a spring. Cloth is modeled as a grid of particles. Each particle is connected to its eight neighbors by a spring, i.e. a stretching constraint. Additionally, for particles that are two-steps away from each other, a bending constraint is used to model the resistance against bending deformation. Additional constraints for modeling self-collision are applied. We refer the readers to~\cite{muller2007position} for more details. 




\section{SoftGym}
To advance research in reinforcement learning in complex environments with an inherently high dimensional state, we propose SoftGym. SoftGym includes a set of tasks related to manipulating deformable objects including rope, cloth, and fluids. As a result, the underlying state representation for the dynamics has a dimension ranging from hundreds to thousands, depending on the number of particles that are used. 

SoftGym consists of three parts: SoftGym-Medium, SoftGym-Hard and SoftGym-Robot, visualized in  Figure~\ref{fig:task_visual}. SoftGym-Medium includes six tasks where we provide extensive benchmarking results. Four more challenging tasks are included in SoftGym-Hard. SoftGym-Medium and SoftGym-Hard using an abstract action space while SoftGym-Robot includes tasks with a Sawyer or Franka robot as the action space.~\footnote{SoftGym-Robot will only be released after obtaining permission from Nvidia.} We describe the details of the action space below.

\subsection{Action Space}
We aim to decouple the challenges in learning low-level grasping skills from high-level planning. As such, we employ abstract action spaces for tasks in SoftGym-Medium and SoftGym-Hard. For rope and cloth manipulation, we use pickers, which are simplifications of a robot gripper and are modeled as spheres which can move freely in the space. A picker can be activated, in which case if it is close to any object, the particle on the object that is the closest to the picker will be attached to the picker and moves with it. More specifically, the action of the agent is a vector of length $4n$, where $n$ is the number of the pickers. For each picker, the agent outputs $(d_x, d_y, d_z, p)$, where $d_x, d_y, d_z$ determine the movement of the picker and $p$ determines whether the picker is activated (picking cloth, $p \geq 0.5$) or deactivated (not picking cloth, $p < 0.5$). For fluid related tasks, we directly actuate the cup holding the fluid. This action space is designed to enable the user to focus on the challenges of high-level planning and to abstract away the low-level manipulation.


Still, in order to reflect the challenges of robotic manipulation, we also provide SoftGym-Robot, where either a Sawyer robot or a Franka robot is used to manipulate objects in the environment (Figure~\ref{fig:task_visual}, bottom-right). Cartesian control of the end-effector is used.

\subsection{Tasks} 
SoftGym-Medium includes six tasks (see Appendix for more details):

\textbf{TransportWater} Move a cup of water to a target position as fast as possible without spilling out the water. The movement of the cup is restricted in a straight line; thus the action of the agent is just a scalar $a = d_x$ that specifies the displacement of the cup. The reward of this task is the negative distance to the target position, with a penalty on the fraction of water that is spilled out. 

\textbf{PourWater} Pour a cup of water into a target cup. The agent directly controls the position and rotation of the cup at hand. Specifically, the action space of the agent is a vector $a = (d_x, d_y, d_{\theta})$, representing the displacement of the cup in the $x,y$ dimension and its rotation around its geometric center. The reward is the fraction of water that is successfully poured into the target cup. 

\textbf{StraightenRope} Straighten a rope starting from a random configuration. The agent controls two pickers. 
The reward is the negative absolute difference between the current distance of the two endpoints of the rope, and the rope length when it is straightened.

\textbf{SpreadCloth} Spread a crumpled cloth on the floor. The agent controls two pickers to spread the cloth. The reward for this task is the area covered by the cloth when viewed top-down.

\textbf{FoldCloth} Fold a piece of flattened cloth in half. The agent controls two pickers. The reward for this task is the distance of the corresponding particles between the two halves of the cloth; we also add a penalty based on the displacement of the cloth from its original position, i.e., we do not want to agent to drag the cloth too far away while folding it.

\textbf{DropCloth} 
This task begins with the agent holding two corners of a piece of cloth with two pickers in the air, and the goal is to lay the cloth flat on the floor. The action space of the agent is the same as that in SpreadCloth, with the additional constraint that the pickers cannot move too close to the floor (i.e. below a given height threshold). As such, a swinging and dropping motion is required to perform the task; dropping the cloth without swinging will result in the cloth being crumpled. The reward of this task is the mean particle-wise $L_2$ distance between the current cloth and a target cloth configuration flattened on the floor. 
  

SoftGym-Hard contains four more tasks:

\textbf{PourWaterAmount} This task is similar to PourWater but requires a specific amount of water poured into the target cup, indicated either in the state representation and marked by a red line in the visual observation.

\textbf{FoldCrumpledCloth} This task is similar to FoldCloth but the cloth is initially crumpled. Thus, the agent may need to spread and fold the cloth at the same time.

\textbf{DropFoldCloth} This task has the same initial state as DropCloth but requires the agent to fold the cloth instead of just laying it on the ground.

\textbf{RopeConfiguration} This task is similar to StraightenCloth but the agent needs to manipulate the rope into a specific configuration from different starting locations. Different goal configurations in the shape of letters of the alphabet can be specified. The reward is computed by finding the minimum bipartite matching distance between the current particle positions and the goal particle positions~\cite{west1996introduction}.

FleX uses a GPU for accelerating simulation. On a Nvidia 2080Ti GPU, all SoftGym tasks run about 4x faster than real time, with rendering. One million simulation steps takes 6 hours (wall-clock time) and corresponds to at least 35 hours for a real robot to collect. More details on the environments, including the variations for each task, can be found in the Appendix.

\section{Methods Evaluated}
We benchmark a few representative policy search algorithms on the tasks in SoftGym.  We group these algorithms into categories which make different assumptions regarding  knowledge about the underlying dynamics or position of particles in the environments.  These algorithms allow us to analyze different aspects of the challenges in learning to manipulate deformable objects. We use the ground-truth reward function for all the baselines, although such rewards are not readily available outside of the simulation environment; computing rewards from high-dimensional observations is an additional challenge that is outside the scope of this paper. We refer to \cite{nair2018visual,pmlr-v80-srinivas18b,florensa2019self,lin2019reinforcement} for recent works towards this challenge.

\subsection{Dynamics Oracle}
The Dynamics Oracle has access to the ground-truth positions and velocities of all particles in the deformable objects, as well as access to the ground-truth dynamics model.  This information is only accessible in simulation. Given this information, we can use gradient free optimization to maximize the return. In this category we benchmark the cross entropy method (CEM)~\cite{rubinstein1999cross} as a representative random shooting algorithm. 
CEM optimizes over action sequences to find the trajectory with the highest return, given the action sequence and the ground-truth dynamics model. Because the ground-truth dynamics model is used, no training is required (i.e.~no parameters are learned) for this method. We use model predictive control~(MPC) for executing the planned trajectories. This baseline shows what can be achieved from a trajectory optimization approach given ground-truth particle positions, velocities, and dynamics; we would expect that methods that only have access to visual observations would have worse performance. However, the performance of the Dynamics Oracle may still be limited due to the exploration strategy employed by CEM.


\subsection{State Oracle}\label{sec:hand-designed features}
Many robotic systems follow the paradigm of first performing state estimation and then using the estimated state as input to a policy. Deformable objects present a unique challenge for manipulation where the dynamical system has a high dimensional state representation, i.e. the position of all particles in the deformable objects. For such high dimensional systems, state estimation becomes harder; furthermore, even assuming perfect state estimation, the high dimensional state space is challenging for any reinforcement learning agent. We explore two state-based methods to explore these challenges.

\textbf{Full State Oracle}  
This method has access to the ground-truth positions of all of the particles in the target object as well as any proprioception information of the robot or the picker, but it does not have access to the ground-truth dynamics. We use these positions as input to a policy trained using SAC~\cite{pmlr-v80-haarnoja18b}; we use the standard multi-layer perceptron~(MLP) as the architecture for the agent. As the observational dimension can vary, e.g.\ cloths with different sizes can have different numbers of particles, we take the maximum number of particles as the fixed input dimension and pad the observation with 0 when it has fewer dimensions. Generally, we expect this baseline to perform poorly, since concatenating the positions of all particles in a single vector as input forces the network to learn the geometric structure of the particles, which adds significantly to the complexity of the learning problem. Given additional task-specific information about the connectivity among the particles, alternative architectures such as graph neural networks may be better at capturing the structure of the particles~\cite{li2019propagation}.


\textbf{Reduced State Oracle} To avoid the challenges of RL from high-dimensional state spaces, this method uses a hand-defined reduced set of the full state as input to the policy. This baseline uses  the same SAC algorithm as the Full State Oracle to train the policy. Estimating this reduced state from a high-dimensional observation of a deformable object,  such as from an image of crumpled cloth, may be challenging; for this oracle baseline, we assume that such a reduced state is perfectly estimated.  Thus, this baseline provides an upper bound on the performance of a real system, assuming that the reduced state captures all of the relevant information needed for the task. Unlike the Dynamics Oracle, this baseline does not assume access to the ground-truth system dynamics.  


For all of the cloth related environments, the reduced state is the positions of the four corners of the cloth. For the StraightenRope environment, we pick 10 evenly-spaced keypoints on the rope, including the two end points, and use the positions of these key points as the reduced state representation. For TransportWater, the reduced state is the size~(width, length, height) of the cup, the target cup position, and the initial height of the water in the cup.  For PourWater, the reduced state is the sizes of both cups, the position of the target cup, the position and rotation of the controlled cup, and the initial height of the water in the cup. For any environment with pickers, the positions of the pickers are also included as part of the reduced state. We note that the set of reduced state we picked may not be sufficient for performing all of the manipulation tasks. For example, knowing the positions of the four corners of a crumpled cloth is not sufficient to infer the full configuration of the cloth, so some information is lost in this reduced state representation.

\subsection{Image Based Observations}
We also evaluate state-of-the-art RL algorithms that directly operate on high dimensional observations. It is important to evaluate methods that use high dimensional observations as input, since it cannot be assumed that a low dimensional state representation (such as that used by the Reduced State Oracle) can always be accurately inferred.


\xingyu{Recent works~\cite{laskin_srinivas2020curl,kostrikov2020image,laskin_lee2020rad} show evidence that the gap between image-based RL and state-based RL can be closed on a range of tasks with the data augmentation in reinforcement learning. Among these, we benchmark CURL-SAC~\cite{laskin_srinivas2020curl}, which uses a model-free approach with a contrastive loss among randomly cropped images, and DrQ~\cite{kostrikov2020image}, which applies data augmentation and regularization to standard RL.} We also evaluate PlaNet~\cite{hafner2019learning}, which learns a latent state space dynamics model for planning. \xingyu{For SpreadCloth and FoldCloth, we additionally benchmark \citet{wu2019learning}, which learns a pick-and-place policy with model-free RL.}

\begin{figure*}[h]
    \centering
    \includegraphics[width=\textwidth]{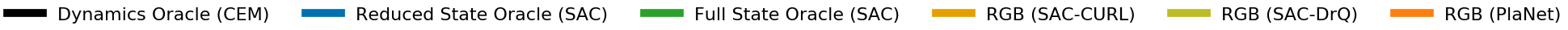} \\
    \includegraphics[width=\textwidth]{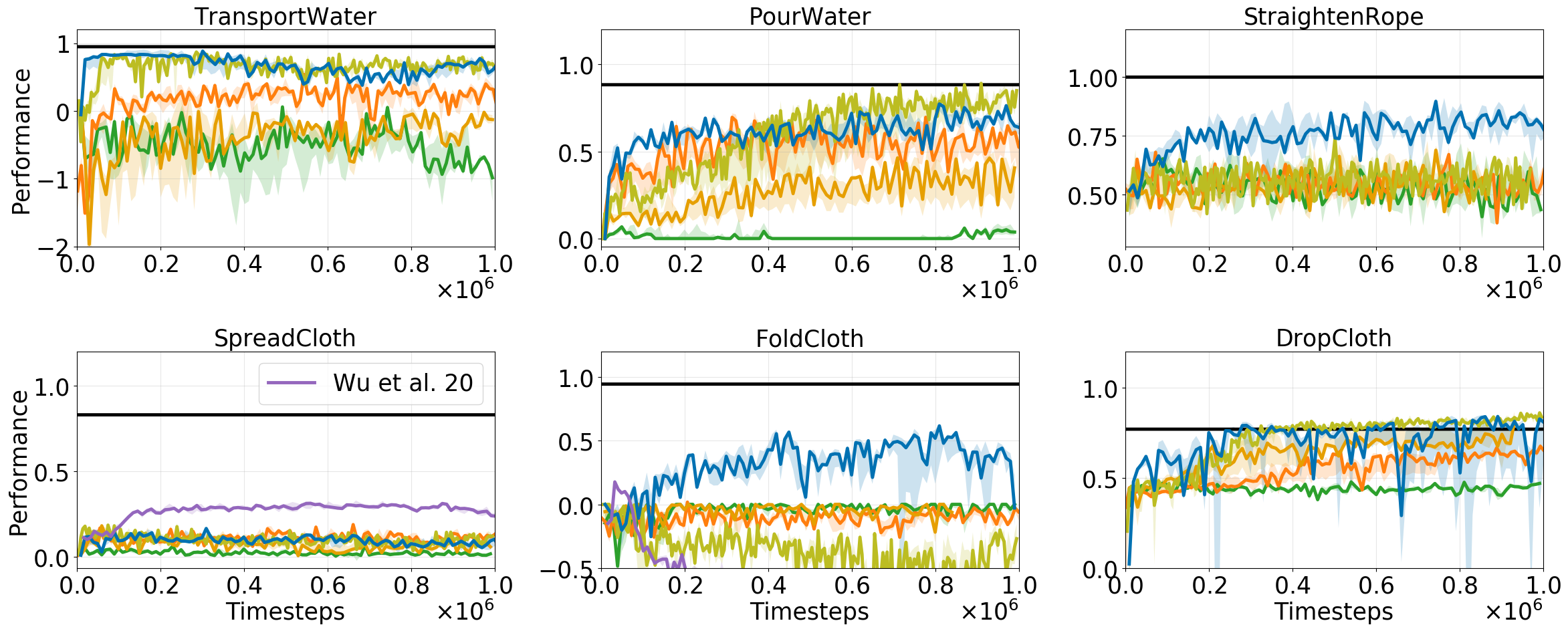} 
    \caption{Normalized performance at the last time step of the episode of all the algorithms on the evaluation set. The x-axis shows the number of training time steps.}
    \label{fig:learning_curve}
\end{figure*}

\section{Experiments}
In this section, we perform experiments with an aim to answer the following questions:
\begin{itemize}
    \item Are SoftGym tasks challenging for current reinforcement learning algorithms?
    \item Is learning with state as sample efficient as learning from high-dimensional observations on SoftGym tasks?
    \item Are the environments realistic enough to reflect the difficulty of learning on a real deformable object dynamical system?
\end{itemize}

\subsection{Experimental Setup}
For each task, we compute a lower bound and upper bound on performance so that we can more easily analyze the performance of each method (see Appendix for details). The lower bound is obtained from a policy that always does nothing. Using these bounds, the performance of each method can then be normalized into $[0, 1]$, although the performance of a policy that performs worse than doing nothing can drop below 0. We run each algorithm for 5 random seeds and plot the median of the normalized performance. Any shaded area in the plots denotes the 25 and 75 percentile. In each task, we pre-sample 1000 variations of the environment. We then separate these task variations into a set of training tasks with 800 variations and a set of evaluation tasks of 200 variations. For CEM, no parameters are trained, so we modify this procedure: instead, we randomly sample 10 task variations from the evaluation set and compute the average performance across the variations. All methods are trained for $10^6$ time steps, except PlaNet, which is trained for $5\times 10^5$ time steps due to computation time.  Please refer to the Appendix for more details of the algorithms and training procedure. Most of the experiments are run on an Nvidia 2080Ti GPU, with 4 virtual CPUs and 40G RAM.

\subsection{Benchmarking results on SoftGym-Medium}
A summary of the final normalized performance of all baselines on the evaluation set is shown in Figure~\ref{fig:learning_curve}. As expected, the Dynamics Oracle performs the best and is able to solve most of the tasks. As the dynamics and ground-truth position of all particles are usually unknown in the real world, this method serves as an upper bound on the performance.

The Reduced State Oracle performs well on tasks where the reduced state captures the task relevant information, such as StraightenRope, TransportWater, FoldCloth, and performs poorly on the tasks which may require more information beyond the reduced state, such as in SpreadCloth, where the positions of the four cloth corners are not sufficient to reason about the configuration of the cloth. We note that the reduced states can be hard to obtain in the real world, such as in TransportWater, where the reduced state includes the amount of the water in the cup. 

More interestingly, we also examine the performance of methods that assume that the agent only has access to image observations.  A robot in the real world will not have access to  ground-truth state information and must use these high dimensional observations as inputs to its policy. We observe that the performance of image-based reinforcement learning (PlaNet, SAC-CURL, or SAC-DrQ) is far below the optimal performance on many tasks.  This is especially true for StraightenRope, SpreadCloth, and FoldCloth, and the learning curves for these tasks seem to imply that even with more training time, performance would still not improve. These methods also have a performance far below the upper bound of 1 on the other tasks (TransportWater, PourWater, DropCloth).  Thus, this evaluation points to a clear need for new methods development for image-based robot manipulation of deformable objects. Compared to the reduced state oracle, image based methods have much worse performance in certain tasks such as FoldCloth or StraightenRope, indicating that there is still a gap between learning from high dimensional observation and learning from state representation.


The Full State Oracle, which uses the position of all particles in the object as input to a policy, performs poorly on all tasks.  This further demonstrates the challenges for current RL methods in learning to manipulate deformable objects which have a variable size and high dimensional state representation. 

For the SpreadCloth task, we additionally compare to previous work~\cite{wu2019learning} that learns a model-free agent for spreading the cloth from image observation. A pick-and-place action space is used here. During training, for collecting data, a random location on the cloth is selected based on a segmentation map and the agent learns to select a place location. The picker will then move to the pick location above the cloth, pick up the cloth, move to the place location and then drop the cloth. In Figure~\ref{fig:learning_curve}, we show the final performance of this method with 20 pick-and-place steps for each episode. While it outperforms the rest of the baselines due to the use of the segmentation map and a better action space for exploration, the result shows that there still exists a large room for improvement. On the other hand, this method does not perform very well on the FoldCloth task.

\subsection{Difficult Future Prediction} 
\begin{figure}[h]
    \centering
    \includegraphics[width=\linewidth]{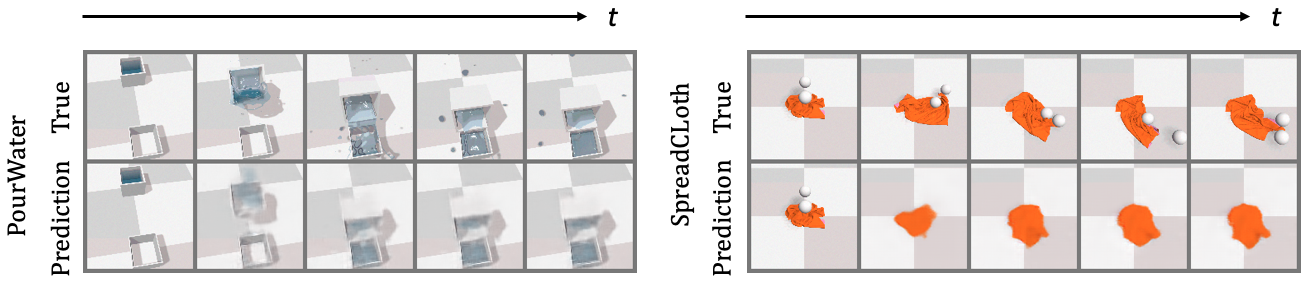}
    \caption{Bottom row: Open-loop prediction of PlaNet. Given an initial set of five frames, PlaNet predicts the following 30 frames. Here we show the last observed frame in the first column and four evenly spaced key frames out of the 30 predicted frames in the last four columns. Top row: Ground-truth future observations.}
    \label{fig:planet_open_loop}
\end{figure}

Why is learning with deformable objects challenging? 
Since PlaNet~\cite{hafner2019learning} learns an autoencoder and a dynamics model, we can visualize the future predicted observations of a trained PlaNet model. We input to the trained model the current frame and the planned action sequence and visualize the open-loop prediction of the future observation. Figure~\ref{fig:planet_open_loop} shows that PlaNet fails to predict the spilled water or the shape of the cloth in deformable manipulation tasks. This provides evidence that since deformable objects have  complex visual observations and dynamics, learning their dynamics is difficult.


\begin{wrapfigure}{R}{8cm}
\centering
\vspace{-12mm}
\includegraphics[width=\linewidth]{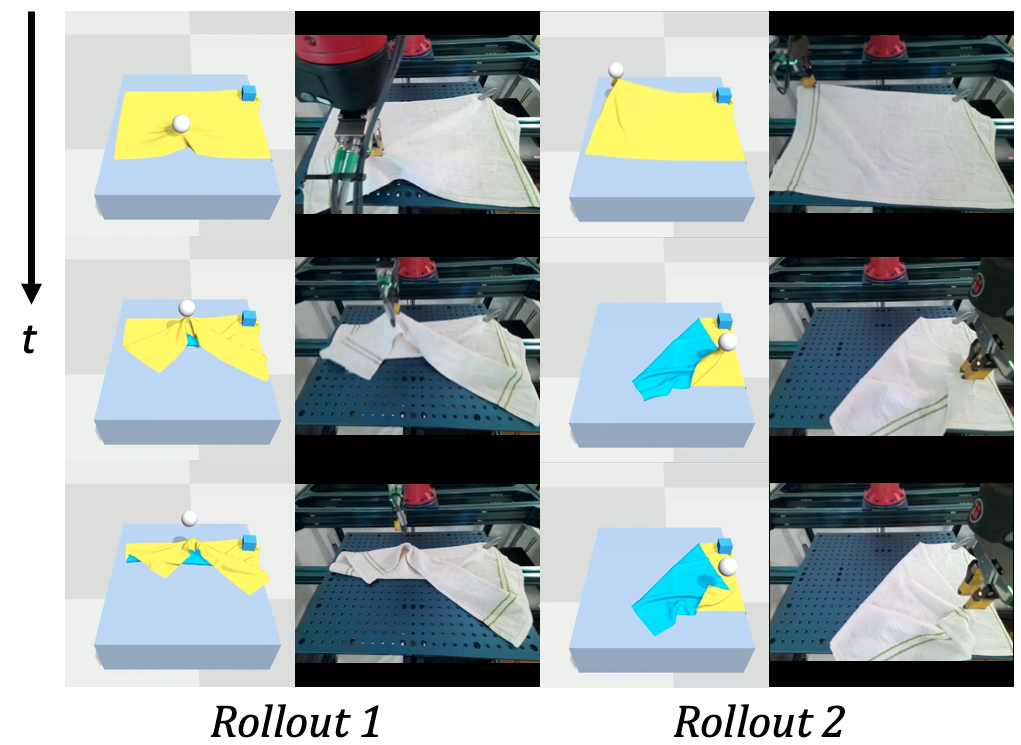}
\caption{Two pick-and-place rollouts both in simulation and in the real world for a cloth manipulation task. For each rollout, the left column shows the simulation; the right shows the real world.}
\label{fig:realism}
\end{wrapfigure}
\subsection{Reality Gap}
Do SoftGym environments reflect the challenges in real world manipulation of deformable objects? Here we take the cloth environments as an example and show that both our cloth modeling and the picker abstraction can be transferred to the real world. We set up a real world cloth manipulation environment with a Sawyer robot with a Weiss gripper, as shown in Figure~\ref{fig:realism}. We perform a series of pick and place actions both in simulation and on the real robot. We can see that the simulated cloth shows similar behaviour to the real one. This demonstration suggests that the simulation environment can reflect the complex dynamics in the real world and that algorithmic improvements of methods developed in SoftGym are likely to correspond to improvements for methods trained in the real world; however, direct sim2real transfer of learned policies is still expected to present a challenge.



\section{Conclusion}
In this paper, we present SoftGym, a set of benchmark environments for deformable object manipulation. We show that manipulating deformable objects presents great challenges in learning from high dimensional observations and complex dynamics. We believe that our benchmark presents challenges to current reinforcement learning algorithms; further, our benchmark should help to ease the comparison of different approaches for deformable object manipulation, which should assist with algorithmic development in this area.



\acknowledgments{This material is based upon work supported by the United States Air Force and DARPA under Contract No. FA8750-18-C-0092, the National Science Foundation under Grant No. IIS-1849154 and LG Electronics. We also thank Sujay Bajrachaya for the help with running the robot experiments in Figure 4.}


\bibliography{softgym}  

\clearpage
\newpage
\appendix

\renewcommand{\figurename}{Supplementary Figure}
\newcommand{\randint}[2]{\text{randint}(#1 #2)}
\newcommand{\unif}[2]{\text{Unif}[#1 #2]}



\section{Environment Details}
\subsection{Observation Space}
Each task supports three types of observation space: Full state of the particles, reduced states and image based observation. For image-based observation, the agent receives an RGB image of the environment rendered by the Flex simulator, with a size $d \times d \times 3$, where $d$ is a controllable parameter. For all our image-based experiments we choose $d = 128$. For the full state observation, the state is the positions of all particles, as well as any state of the action space or other rigid objects in the scene.

We now detail the reduced state representation for each task in SoftGym.

\textbf{TransportWater}: the reduced states are the size~(width, length, height) of the cup, the target cup position, height of the water in the cup, amount of water inside and outside of the cup.

\textbf{PourWater and PourWaterAmount}: the reduced states are the sizes of both cups, the $x,y$-position and rotation of the controlled cup, the initial distance between the controlled cup and the current cup, the height of the water in the cup and the amount of water in both cups. For PourWaterAmount, we have an additional value indicating the amount of water to be poured.

\textbf{Rope Environments}: For rope enviornments, including the StraightenRope and RopeConfiguration, we pick 10 evenly-spaced keypoints on the rope, including the two end points, and use the positions of these key points as the reduced state.

\textbf{Cloth Environments}: For all of the cloth related environments~(SpreadCloth, FoldCloth(Crumpled), DropCloth, DropFoldCltoh), the reduced states are the positions of the four corners of the cloth. 

For environments using any pickers or robots, the positions of the pickers or joint positions of the robot are included in the reduced state.

\subsection{Action Space}
For all environments, we normalize the action space to be within $[-1, 1]$ for the agents. Below we will describe the un-normalized action range for each environment, using meter or radian as the unit by default.

\textbf{TransportWater:} The motion of the cup is constrained to be in one dimension. The action is also in one dimension and is the increment of the the position of the cup along the dimension. The action range is $[-0.011, 0.011]$.

\textbf{PourWater, PourWaterAmount:} The action is $a = (dx, dy, d\theta)$, denoting the change of the position and rotation of the cup. $dx, dy \in[-0.01, 0.01]$ and $d\theta \in [-0.015, 0.015]$. 

\textbf{Picker:} For the cloth and rope environments, we use either two pickers or one robot. A picker abstracts a controller which can pick and place objects. A picker is modeled as a sphere with a radius of $0.05$. For each picker, the action is $a=(dx, dy, dz, d)$. $dx, dy, dz \in [-0.01, 0.01]$ indicate the change of the position of the picker. $d \in [0, 1]$ indicates the picking state of the picker. If $d>0.5$, the particle closest to the picker will be attached to the picker and follow its movement. When $d<0.5$, the picked particle will be released.

\subsection{Task Variations}\label{sec:task_variations}
\begin{table*}[h]
    \centering
    \small
    \begin{tabular}{cc}
    \toprule
    Environment & Variations  \\ \hline
    TransportWater & cup size, water volume \\
    PourWater & size of both cups, target cup position, water volume \\
    PourWaterAmount & size of both cups, target cup position, water volume, goal volume \\
    StraightenRope, RopeConfiguration & Initial rope configuration \\
    SpreadCloth, FoldClothCrumpled & Cloth size, initial configuration \\
    FoldCloth & Cloth size, random rotation of the inital configuration \\
    DropCloth, DropFoldCloth & Cloth size, initial height \\
    
    \bottomrule
    \end{tabular}
    \caption{Different task variations in all tasks. Refer to the appendix for more details of the ranges of the variations and how they are generated.}
    \label{tab:task_variation}
\end{table*}
\begin{figure}[h]
    \centering
    \includegraphics[width=\textwidth]{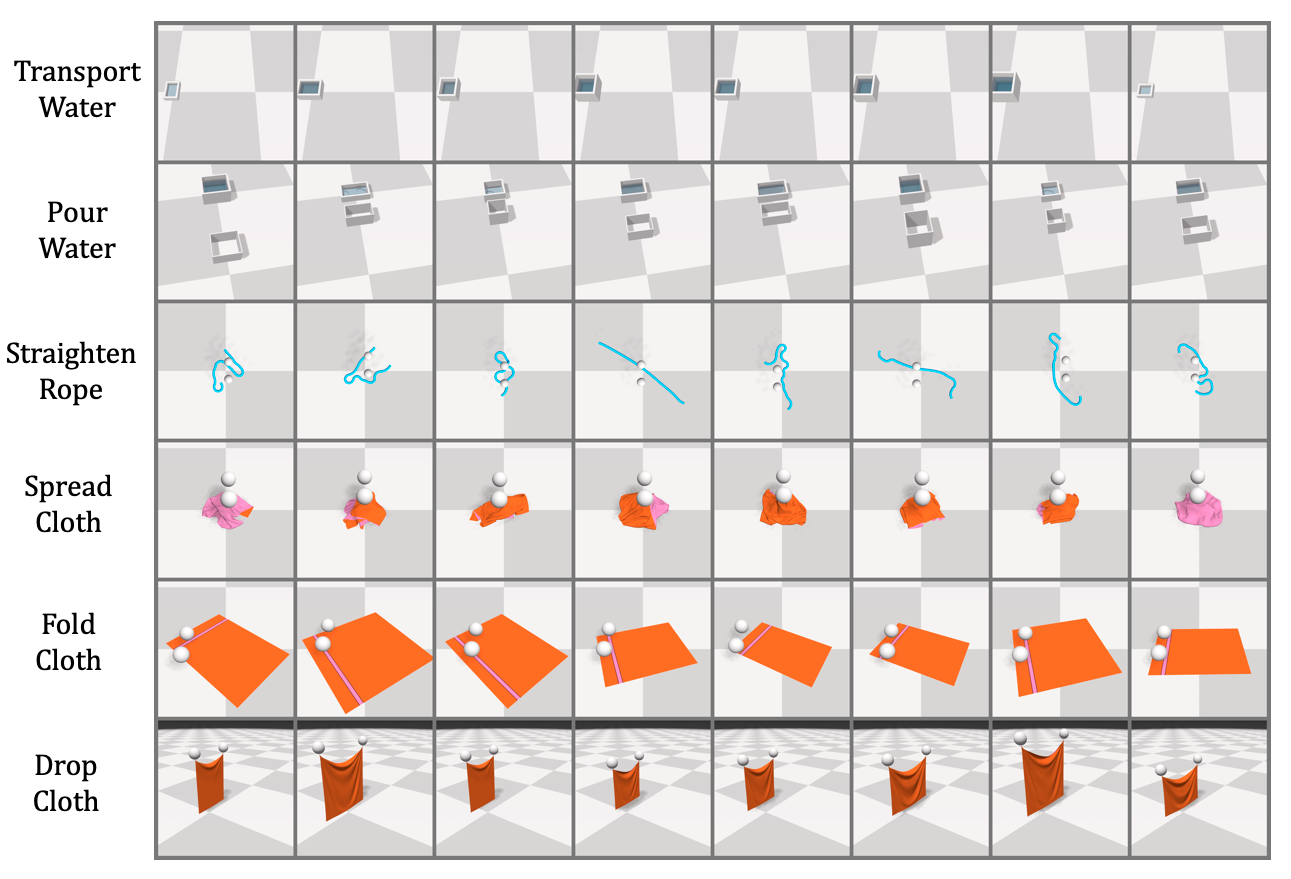}
    \caption{Illustration of task variations. Each image shows the task after the initial reset. Variations of tasks in SoftGym-Hard are omitted here due to similarity to the ones shown.}
    \label{fig:task_variation}
\end{figure}
In this section we detail how we generate the task variations. 
Supplementary Figure \ref{fig:task_variation} shows some of the task variations.
Most of the practical tasks related to deformable object manipulation, such as laundry folding, require the agent to deal with variations of the objects in the environment, such as the size, shape, and physical properties. We summarize the variations of each task that we include in SoftGym in Table \ref{tab:task_variation}. 

\textbf{PourWater, PourWaterAmount:}
For this task, both the controlled cup and the target cup are modeled as cuboids without the top face. We vary the height, length, and width of both cups, the distance between the controlled cup and the target cup, as well as the volume of water in the controlled cup.
The water is initially generated in a shape of cuboid. Denote the number of particles along each dimension of the water cuboid as $l_w, w_w, h_w$ respectively. We vary the width $w_w$ and the length $l_w$ of the water cuboid in the range $[4, 13]$. For the height $h_w$, we first randomly select a water level between 'medium' and 'large'.
Let $m = \min(w_w, l_w)$. For level 'medium', the height is $h_w = \lfloor 3.5m \rfloor$. For level `large', the height is $h_w = 4m$.  The total number of water particles is $v = w_w \cdot h_w \cdot l_w$.

Given the volume of water, we then create a cup for holding that amount of water. Denote the radius of the water particles as $r = 0.033$. The width and length of the controlled cup is $w_{cc} = w_w \cdot r + 0.1$ and $l_{cc} = l_w \cdot r + 0.1$, and the width and length of the target cup is $w_{tc} = w_w \cdot r + 0.07$ and $l_{tc} = l_w \cdot r + 0.07$. 
Let $h = v / ((w_w + 1)(l_w + 1))$ be the number of particles in the height of the water cuboid.
For medium volume of water, we have $h_{cc} = h \cdot r / 2 + 0.001 \cdot \unif{-0.5, 0.5}$. For large volume of water, we have $h_{cc} = h \cdot r / 3 + 0.001 \cdot \unif{0, 1}$. The height of the target cup is simply computed as $h_{tc} = h_{cc} + \unif{0, 0.1}$.

The distance between the controlled cup and target cup is sampled  $m \cdot \unif{0.05m, 0.09m} + (w_w + 4) r / 2$.

For PourWaterAmount, the goal volume is sampled from $0.1 + \text{Unif}[0, 1]$$ \times 0.9$.

\textbf{TransportWater:} We vary the volume of water and size of cup in this task. The variation is generated almost exactly the same as in PourWater, with the following exceptions. For the medium volume of water, the cup height is computed as $h_{cc} = h \cdot r / 2$, and for the large volume of water, the cup height is computed as $h_{cc} = h \cdot r / 3 + 0.0015m$.

\textbf{SpreadCloth, FoldClothCrumpled:} We vary the size of the cloth. The cloth is modeled as a particle grid with width $w$ and length $l$ (the number of particles). We sample $w$ and $l$ from $\randint{60, 120}$. We also vary the initial crumpled shape of the cloth. This is done by first randomly picking a particle on the cloth, lifting it up to a random height sampled from $\unif{0, 0.5}$, and then dropping it.

\textbf{FoldCloth:} In this task we only vary the size of the cloth. Similar to the SpreadCloth case, the width of the cloth $w$ is sampled from $\randint{60, 120}$. The initial state of the cloth is always flattened and centered at the origin.

\textbf{DropCloth, DropFoldCloth:} In this task we vary the size of the cloth in the same way in SpreadCloth. We lift the cloth up by picking up its two corners.

\textbf{StraightenRope:} In this task we use a rope with a fixed length and only vary its initial twisted shape. We generate different twisted shapes by randomly choosing a particle on the rope, picking it up, and then dropping it. We repeat this process for 4 times to make sure the generated shapes are different. 

\subsection{Training and Evaluation}
For computation efficiency, we pre-compute 1000 task variations and their initial states for each environment. Out of the 1000 task variations, 800 variations are used during training and 200 variations are used for evaluation.

\textbf{Performance Metric} Besides the reward, we compute a performance metric for each task at each time step, which is the same as the reward without any scaling. At the beginning of each episode, we compute an upper-bound and a lower-bound for the performance metric. For example, for SpreadCloth task, the performance is the covered area of the cloth and the upper-bound is when the cloth is flattened. For any task where the performance is a negative distance function, its upper-bound would be zero. For all tasks except StraightenRope, the lower-bound is the performance achieved at the first time step, which corresponds to the achieved performance when the policy does nothing. For StraightenRope, the lower bound is the possible minimal reward, which is the negative value of the straightened rope's length. Given the upper-bound and lower-bound $u, l$, we normalize the performance at each time step by $$\hat{s} = \frac{s-l}{u-l},$$ where $\hat{s}$ is the normalized performance. The normalized performnace at the last time step is reported throughout the paper unless explicitly specified.

\section{Algorithm Details}
For all the tasks and algorithms, we use a discounting factor of $\gamma = 0.99$ when it applies. The action repetition and task horizon are summarized in table \ref{tab:env_params}.

\begin{table}[h]
    \centering
    \begin{tabular}{lllllll}
    \toprule
      Parameter  & \tabincell{l}{Transport\\Water} & \tabincell{l}{Pour\\Water} & \tabincell{l}{Straighten\\Rope} &
      \tabincell{l}{Spread\\Cloth} & \tabincell{l}{Fold\\Cloth} & \tabincell{l}{Drop\\Cloth}   \\ \hline
    Action Repetition & 8 & 8 & 8 & 8 & 8 & 32  \\
    Task Horizon    & 75 & 100 &75 & 100 & 100 & 15  \\
\bottomrule
    \end{tabular}
    \caption{Action repetition and task horizon.}
    \label{tab:env_params}
\end{table}

\subsection{CEM with Dynamics Oracle}
For CEM, we use 10 optimization iteration. Model predictive control is used. Different planning horizon is used for different environments, as summarized in Table~\ref{tab:cem_planning_horizon}. A total of 21K environment steps are used for making each decision. The number of candidate trajectories during each planning is thus $21K/planning\_horizon$. The top $10\%$ candidates are selected as the elites for fitting the posterier distribution within each optimization iteration.

\begin{table}[h]
    \centering
    \begin{tabular}{lllllll}
    \toprule
      Parameter  & \tabincell{l}{Transport\\Water} & \tabincell{l}{Pour\\Water} & \tabincell{l}{Straighten\\Rope} &
      \tabincell{l}{Spread\\Cloth} & \tabincell{l}{Fold\\Cloth} & \tabincell{l}{Drop\\Cloth}   \\ \hline
     Planning Horizon    & 7 & 40 & 15 & 15 & 30 & 15  \\
\bottomrule
    \end{tabular}
    \caption{Task specific planning horizon for CEM}
    \label{tab:cem_planning_horizon}
\end{table}

\subsection{SAC and CURL-SAC}
We use the CURL-SAC implementation from the released code\footnote{\url{https://github.com/MishaLaskin/curl}}. Both Q-value network and the policy network are MLPs with 2 hidden layers of 1024 neurons with ReLU as activation function. The hyper-parameters of SAC are summarized in Table~\ref{tab:SACTD3_params}. To achieve learning stability, we tuned the reward scaling and learning rate for both SAC and CURL-SAC, for each environment. The parameters are summarized in Table~\ref{tab:sac_lr}.

\begin{table}[h]
    \centering
    \begin{tabular}{ll}
    \toprule
       Parameter & SAC  \\ \hline
       batch size & 128     \\
       initial steps & 1000 \\
       replay buffer size  &1e5 \\
       target smoothing coefficient &0.01\\ 
       alpha & automatic tuning \\
       delayed policy update period & 2 \\ 
       target update interval & 2 \\
\bottomrule
    \end{tabular}
    \caption{General hyper-parameters for SAC.}
    \label{tab:SACTD3_params}
\end{table}

\begin{table}[h]
    \centering
    \begin{tabular}{lllllll}
    \toprule
      Parameter  & \tabincell{l}{Transport\\Water} & \tabincell{l}{Pour\\Water} & \tabincell{l}{Straighten\\Rope} & \tabincell{l}{Spread\\Cloth} & \tabincell{l}{Fold\\Cloth} & \tabincell{l}{Drop\\Cloth}   \\ \hline
      \textit{Reduced State} & \\
      \hspace{5mm}\tabincell{c}{learning rate} & 1e-3 & 1e-3 & 1e-3 & 1e-3 & 5e-4 & 1e-3  \\
      \hspace{5mm}\tabincell{c}{reward scaling} & 20 & 20 & 50 & 50 & 50 & 50  \\
      \hspace{5mm}\tabincell{c}{learning rate decay} & - & - & yes & - & - & -  \\
      \midrule
      \textit{Image} & \\
      \hspace{5mm}\tabincell{c}{learning rate} & 3e-4 & 3e-4 & 3e-4 & 3e-4 & 1e-4 & 3e-4  \\
      \hspace{5mm}\tabincell{c}{learning rate decay} & - & - & yes & yes  & - & -  \\
      \hspace{5mm}\tabincell{c}{reward scaling} & 20 & 20 & 50 & 50 & 50 & 50  \\
\bottomrule
    \end{tabular}
    \caption{SAC task dependent hyper-parameters. If learning rate decay is applied, the actor learning rate is halved every 75K steps and the critic learning rate is halved every 100K steps.}
    \label{tab:sac_lr}
\end{table}

\subsection{DrQ}
We use the author released code\footnote{\url{https://github.com/denisyarats/drq}} for the benchmarking with mostly default hyper-parameters. The only change in the hyper-parameter is that we use images of size $128 \times 128$ instead of $84 \times 84$ as in the released code, so we change the padding of the image from $4$ to $6$. We also tune the reward scaling parameter for different tasks, as summarized  in Table~\ref{tab:sac_lr}.

\begin{table*}[h!]\centering
\begin{tabular}{@{}lp{70mm}}
\toprule
Parameter & Value\\
\midrule
\textit{training} & \\
\hspace{5mm}optimizer & Adam \cite{kingma2014adam} \\
\hspace{5mm}learning rate & 0.001 \\
\hspace{5mm}Adam $\epsilon$ & 0.0001 \\
\hspace{5mm}experience replay size & $10^6$ \\
\hspace{5mm}explore noise & 0.3 \\
\hspace{5mm}batch size & 50 \\
\hspace{5mm}dynamics chunk size & 50 \\
\hspace{5mm}free nats & 3 \\

\midrule
\textit{CEM planning} & \\
\hspace{5mm}planning horizon & 24 \\
\hspace{5mm}optimization iteration & 10\\
\hspace{5mm}candidate samples & 1000 \\
\hspace{5mm}top candidate & 100\\
\bottomrule
\end{tabular}
\caption{Hyper-parameters for PlaNet}
\label{tab:planet_params}
\end{table*}

\subsection{PlaNet}
\begin{table}
    \centering
    \begin{tabular}{ccccc}
    \toprule
    layer & input channel & output channel & kernel size & stride \\ \hline
    1     & 1024 &  128 & 5 & 2 \\
    2     & 128 &  64 & 5 & 2 \\
    3     & 64 &  32 & 5 & 2 \\
    4     & 32 &  16 & 6 & 2 \\
    5  & 16 &  3 & 6 & 2 \\
    \bottomrule
    \end{tabular}
    \caption{Architecture of the deconvolutional neural network (VAE decoder) in PlaNet.}
    \label{tab:DCNN}
\end{table}
\begin{table}
    \centering
    \begin{tabular}{ccccc}
    \toprule
    layer & input channel & output channel & kernel size & stride \\ \hline
    1     & 3 &  16 & 4 & 2 \\
    2     & 16 &  32 & 4 & 2 \\
    3     & 32 &  64 & 4 & 2 \\
    4     & 64 &  128 & 4 & 2 \\
    5  & 128 &  256 & 4 & 2 \\ 
    \bottomrule
    \end{tabular}
    \caption{Architecture of the encoding CNN.}
    \label{tab:encodingCNN}
\end{table}
PlaNet takes the image observation as input. The image is first processed by a convolutional neural network to produce an embedding vector. The architecture of the encoding CNN is shown in table \ref{tab:encodingCNN}. After the final convolution layer, the extracted features are flattened to be a vector, then transformed to an embedding vector by a linear layer. Different algorithms use different sizes for the embedding vector. For PlaNet and RIG, the size is 1024; For SAC and TD3, the size is 256. Different algorithms then process this embedding vector in different ways. 

We use a GRU~\cite{cho2014learning} with 200 hidden nodes as the deterministic path in the dynamics model. All functions are implemented as a two-layer MLP with 200 nodes for each layer and ReLU as the activation function. We refer to~\cite{hafner2018learning} for more details. We do not include the latent over-shooting in our experiment as it does not improve much over the one-step case.

During training, we first collect 5 episodes from a random policy to warm-up the replay buffer. Then, for each training epoch, we first store 900 time steps of experiences collected from the current planning agent and perform 100 gradient updates. The full hyper-parameters are listed in Table~\ref{tab:planet_params}.

On an Nvidia 2080Ti GPU with 4 virtual CPUs and 40G RAM, training PlaNet for 1M steps takes around 120 hours.

\subsection{Wu et al. 20}
For the SpreadCloth and FoldCLoth task, we additionally compare to previous work~\cite{wu2019learning} that learns a model-free agent for spreading the cloth from image observation. We take the official implementation from the authors~\footnote{\url{https://github.com/wilson1yan/rlpyt}}. Here, the action space is pick-and-place. Followed the approach in the paper, during exploration, a random point on the cloth is selected (with a heuristic method for cloth segmentation). The picker then goes to the picked location, picks up the cloth and moves to a place location given by the agent, waits for 20 steps and then drops the cloth. Default hyper-parameters in the original code are used.

\section{CEM with Different Planning Horizons}
\begin{figure}[h]
    \centering
    \includegraphics[width=0.8\textwidth]{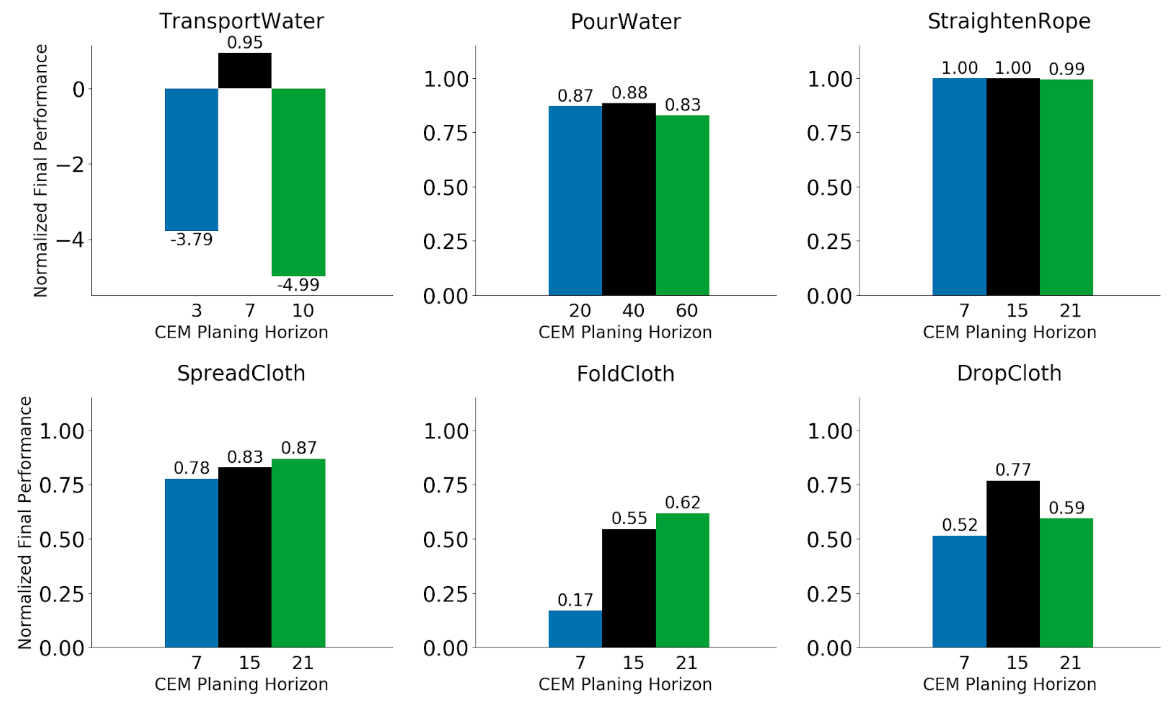}
    \caption{Performance of CEM with different planning horizons for each task.}
    \label{fig:cem_different_horizon}
\end{figure}

We additionally evaluate CEM with different planning horizons for each task. The results are shown in Figure~\ref{fig:cem_different_horizon}. We see that the performance of CEM is sensitive to the planning horizon in TransportWater, FoldCloth and DropCloth, whereas the performance is relatively stable in the other tasks. The black bar is the performance that we report in the main paper.

\end{document}